# TGIF: A New Dataset and Benchmark on Animated GIF Description

Yuncheng Li  
University of Rochester

Yale Song  
Yahoo Research

Liangliang Cao  
Yahoo Research

Joel Tetreault  
Yahoo Research

Larry Goldberg  
Yahoo Research

Alejandro Jaimes  
AiCure

Jiebo Luo  
University of Rochester

## Abstract

*With the recent popularity of animated GIFs on social media, there is need for ways to index them with rich metadata. To advance research on animated GIF understanding, we collected a new dataset, Tumblr GIF (TGIF), with 100K animated GIFs from Tumblr and 120K natural language descriptions obtained via crowdsourcing. The motivation for this work is to develop a testbed for image sequence description systems, where the task is to generate natural language descriptions for animated GIFs or video clips. To ensure a high quality dataset, we developed a series of novel quality controls to validate free-form text input from crowdworkers. We show that there is unambiguous association between visual content and natural language descriptions in our dataset, making it an ideal benchmark for the visual content captioning task. We perform extensive statistical analyses to compare our dataset to existing image and video description datasets. Next, we provide baseline results on the animated GIF description task, using three representative techniques: nearest neighbor, statistical machine translation, and recurrent neural networks. Finally, we show that models fine-tuned from our animated GIF description dataset can be helpful for automatic movie description.*

## 1. Introduction

Animated GIFs have quickly risen in popularity over the last few years as they add color to online and mobile communication. Different from other forms of media, GIFs are unique in that they are spontaneous (very short in duration), have a visual storytelling nature (no audio involved), and are primarily generated and shared by online users [3]. Despite its rising popularity and unique visual characteristics, there is a surprising dearth of scholarly work on animated GIFs in the computer vision community.

In an attempt to better understand and organize the growing number of animated GIFs on social media, we constructed an animated GIF description dataset which consists of user-generated animated GIFs and crowdsourced natural language descriptions. There are two major challenges to this work: (1) We need **a large scale dataset** that captures a wide variety of interests from online users who produce animated GIFs; (2) We need **automatic validation methods** that ensure high quality data collection at scale, in order to deal with noisy user-generated content and annotations. While it is difficult to address these two challenges at once, there has been great progress in recent years in collecting large scale datasets in computer vision [33, 20, 34, 31]. Our work contributes to this line of research by collecting a new large scale dataset for animated GIF description, and by presenting automatic validation methods that ensure high quality visual content and crowdsourced annotations.

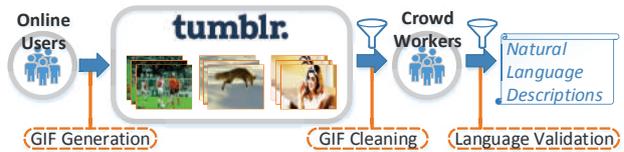

Figure 1: Our TGIF dataset contains 100K animated GIFs and 120K natural language descriptions. (a) Online users create GIFs that convey short and cohesive visual stories, providing us with well-segmented video data. (b) We crawl and filter high quality animated GIFs, and (c) crowdsource natural language descriptions ensuring strong visual/textual association.

Our dataset, *Tumblr GIF* (TGIF), contains 100K animated GIFs collected from Tumblr, and 120K natural language sentences annotated via crowdsourcing. We developed extensive quality control and automatic validation methods for collecting our dataset, ensuring strong and unambiguous association between GIF and sentence. In addition, we carefully evaluate popular approaches for video description and report several findings that suggest future research directions. It is our goal that our dataset and baseline results will serve as useful resources for future video description and animated GIF research.[1]

Our work is in part motivated by the recent work on im-age and video description [40, 9, 14, 21, 38]. Describing animated GIFs, or image sequences in general, is different from the image captioning task (e.g., MS-COCO [20]) because of motion information involved between frames. Recent movie description datasets, such as M-VAD [34] and MPII-MD [31], made the first attempt towards this direction by leveraging professionally annotated descriptive video service (DVS) captions from commercial movies. However, as we show later in this paper, such datasets contain certain characteristics not ideal for image sequence description (i.e., poorly segmented video clips, descriptions with contextual information not available within a provided clip).

---

[1] We use video description, image sequence description, and animated GIF description interchangeably, as they all contain sequences of images.

We make the following contributions in this paper:
1. We collect a dataset for animated GIF description. We solved many challenges involved in data collection, including GIF filtering, language validation and quality control.
2. We compare our dataset to other image and video description datasets, and find that animated GIFs are temporally well segmented and contain cohesive visual stories.
3. We provide baseline results on our dataset using several existing video description techniques. Moreover, we show that models trained on our dataset can improve performance on the task of automatic movie description.
4. We make our code and dataset publicly available at https://github.com/raingo/TGIF-Release

## 2. Related Work

There is growing interest in automatic image and video description [20, 34, 31, 44]. We review existing datasets and some of the most successful techniques in this domain.

**Datasets.** For image captioning, the SBU dataset [25] contains over 1 million captioned images crawled from the web, while the MS-COCO dataset [20] contains 120K images and descriptions annotated via crowdsourcing. The VQA [1] and the Visual Madlibs [44] datasets are released for image captioning and visual question answering.

In the video domain, the YouTube2Text dataset [5, 11] contains 2K video clips and 120K sentences. Although originally introduced for the paraphrasing task [5], this dataset is also suitable for video description [11]. The TACoS dataset [30] contains 9K cooking video clips and 12K descriptions, while the YouCook dataset [8] contains 80 cooking videos and 748 descriptions. More recently, the M-VAD [34] and MPII-MD [31] datasets use the descriptive video service (DVS) from commercial movies, which is originally developed to help people with visual impairment understand non-narrative movie scenes. Since the two datasets have similar characteristics, the Large Scale Movie Description Challenge (LSMDC) makes use of both datasets [34, 31]. Our work contributes to the video domain with 1) animated GIFs, which are well-segmented video clips with cohesive stories, and 2) natural language descriptions with strong visual/textual associations.

**Techniques.** Image and video description has been tackled by using established algorithms [42, 10, 25, 17, 32]. Ordonez *et al*. [25] generate an image caption by finding $k$ nearest neighbor images from 1 million captioned images and summarizing the retrieved captions into one sentence. Rohrbach *et al*. [32] formulate video description as a translation problem and propose a method that combines semantic role labeling and statistical machine translation.

Recent advances in recurrent neural networks has led to end-to-end image and video description techniques [40, 9, 14, 21, 39, 38, 43, 26, 41]. Venugopalan *et al*. [39] represent video by mean-pooling image features from frames, while Li *et al*. [43] apply the soft-attention mechanism to represent each frame of a video, which is then output to an LSTM decoder [12] to generate a natural language description. More recently, Venugopalan *et al*. [38] use an LSTM to encode image sequence dynamics, formulating the problem as sequence-to-sequence prediction. In this paper, we evaluate three representative techniques (nearest neighbor [25], statistical machine translation [32], and LSTMs [38]) and provide benchmark results on our dataset.

### 2.1. Comparison with LSMDC

In essence, movie description and animated GIF description tasks both involve translating image sequence to natural language, so the LSMDC dataset may seem similar to the dataset proposed in this paper. However, there are two major differences. First, our set of animated GIFs was created by online users while the LSMDC was generated from commercial movies. Second, our natural language generations were crowdsourced whereas the LSMDC descriptions were carried out by descriptive video services (DVS). This led to the following differences between the two datasets[2]:

**Language complexity.** Movie descriptions are made by trained professionals, with an emphasis on describing key visual elements. To better serve the target audience of people with visual impairment, the annotators use expressive phrases. However, this level of complexity in language makes the task very challenging. In our dataset, our workers are encouraged to describe major visual content directly, and not to use overly descriptive language. As an example to illustrate the language complexity difference, the LSMDC dataset described a video clip as "amazed someone starts to play the rondo again.", while for the same clip, a crowd worker described as "a man plays piano as a woman stands and two dogs play."

**Visual/textual association.** Movie descriptions often contain contextual information not available within a single movie clip; they sometimes require having access to other parts of a movie that provide contextual information. Our descriptions do not have such issue because each animated

---
[2]Side-by-side comparison examples: https://goo.gl/ZGYIYh

GIF is presented to workers without any surrounding context. Our analysis confirmed this, showing that 20.7% of sentences in LSMDC contain at least two pronouns, while in our TGIF dataset this number is 7%.

**Scene segmentation.** In the LSMDC dataset, video clips are segmented by means of speech alignment, aligning speech recognition results to movie transcripts [31]. This process is error-prone and the errors are particularly harmful to image sequence modeling because a few irrelevant frames either at the beginning or the end of a sequence can significantly alter the sequence representation. In contrast, our GIFs are by nature well segmented because they are carefully curated by online users to create high quality visual content. Our user study confirmed this; we observe that 15% of the LSMDC movie clips v.s. 5% of animated GIFs is rated as not well segmented.

## 3. Animated GIF Description Dataset

### 3.1. Data Collection

We extract a year's worth of GIF posts from Tumblr using the public API[3], and clean up the data with four filters: (1) **Cartoon.** We filter out cartoon content by matching popular animation keywords to user tags. (2) **Static.** We discard GIFs that show little to no motion (basically static images). To detect static GIFs, we manually annotated 7K GIFs as either static or dynamic, and trained a Random Forest classifier based on C3D features [36]. The 5-fold cross validation accuracy for this classifier is 89.4%. (3) **Text.** We filter out GIFs that contain text, e.g., memes, by detecting text regions using the Extremal Regions detector [23] and discarding a GIF if the regions cover more than 2% of the image area. (4) **Dedup.** We compute 64bit DCT image hash using pHash [45] and apply multiple index hashing [24] to perform $k$ nearest neighbor search ($k = 100$) in the Hamming space. A GIF is considered a duplicate if there are more than 10 overlapping frames with other GIFs. On a held-out dataset, the false alarm rate is around 2%.

Finally, we manually validate the resulting GIFs to see whether there is any cartoon, static, and textual content. Each GIF is reviewed by at least two annotators. After these steps, we obtain a corpus of 100K clean animated GIFs.

### 3.2. Data Annotation

We annotated animated GIFs with natural language descriptions using the crowdsourcing service CrowdFlower. We carefully designed our annotation task with various quality control mechanisms to ensure the sentences are both syntactically and semantically of high quality.

A total of 931 workers participated in our annotation task. We allowed workers only from Australia, Canada,

[3] https://www.tumblr.com/docs/en/api/v2

Figure 2: The instructions shown to the crowdworkers.

New Zealand, UK and USA in an effort to collect fluent descriptions from native English speakers. Figure 2 shows the instructions given to the workers. Each task showed 5 animated GIFs and asked the worker to describe each with one sentence. To promote language style diversity, each worker could rate no more than 800 images (0.7% of our corpus). We paid 0.02 USD per sentence; the entire crowdsourcing cost less than 4K USD. We provide details of our annotation task in the supplementary material.

**Syntactic validation.** Since the workers provide free-form text, we automatically validate the sentences before submission. We do the following checks: The sentence (1) contains at least 8, but no more than 25 words (white space separated); (2) contains only ASCII characters; (3) does not contain profanity (checked by keyword matching); (4) should be typed, not copy/pasted (checked by disabling copy/paste on the task page); (5) should contain a main verb (checked by using standard POS tagging [35]); (6) contains no named entities, such as a name of an actor/actress, movie, country (checked by the Named Entity Recognition results from DBpedia spotlight [6]); and (7) is grammatical and free of typographical errors (checked by the LanguageTool[4]).

This validation pipeline ensures sentences are *syntactically* good. But it does not ensure their *semantic* correctness, i.e., there is no guarantee that a sentence accurately describes the corresponding GIF. We therefore designed a semantic validation pipeline, described next.

[4] https://languagetool.org/

**Semantic validation.** Ideally, we would like to validate the semantic correctness of every submitted sentence (as we do for syntactic validation). But doing so is impractical. We turn to the "blacklisting" approach, where we identify workers who underperform and block them accordingly.

We annotated a small number of GIFs and used them to measure the performance of workers. We collected a validation dataset with 100 GIFs and annotated each with 10 sentences using CrowdFlower. We carefully hand-picked the GIFs whose visual story is clear and unambiguous. After collecting the sentences, we manually reviewed and edited them to make sure they meet our standard.

Using the validation dataset, we measured the semantic relatedness of sentence to GIF using METEOR [18], a metric commonly used within the NLP community to measure machine translation quality. We compare a user-provided sentence to 10 reference sentences using the metric, accept a sentence if the METEOR score is above a threshold (empirically set at 20%). This will filter out junk sentences, e.g., "this is a funny GIF taken in a nice day," but retain sentences with similar semantic meaning as the reference sentences.

We used the dataset in both the qualification and the main tasks. In the qualification task, we provided 5 GIFs from the validation dataset and approved a worker if they successfully described at least four tests. In the main task, we randomly mixed one validation question with four main questions; a worker is blacklisted if the overall approval rate on validation questions falls below 80%. Because validation questions are indistinguishable from normal task questions, workers have to continue to maintain a high level of accuracy in order to remain eligible for the task.

As we run the CrowdFlower task, we regularly reviewed failed sentences and, in the case of a false alarm, we manually added the failed sentence to the reference sentence pool and removed the worker from the blacklist. Rashtchian *et al.* [28] and Chen *et al.* [5] used a similar prescreening strategy to approve crowdworkers; our strategy to validate sentences *during* the main task is unique to our work.

## 4. Dataset Analysis

We compare TGIF to four existing image and video description datasets: MS-COCO [20], M-VAD [34], MPII-MD [31], and LSMDC [34, 31].

**Descriptive statistics.** We divide 100K animated GIFs into 90K training and 10K test splits. We collect 1 sentence and 3 sentences per GIF for the training and test data, respectively. Therefore, there are about 120K sentences in our dataset. By comparison, the MS-COCO dataset [20] has 5 sentences and 40 sentences for each training and test sample, respectively. The movie datasets have 1 professionally created sentence for each training and test sample. On average, an animated GIF in our dataset is 3.10 seconds long, a video clip in the M-VAD [34] and the MPII-MD [31]

|     | TGIF    | M-VAD  | MPII-MD | LSMDC   | COCO    |
|-----|---------|--------|---------|---------|---------|
| (a) | **125,781** | 46,523 | 68,375  | 108,470 | 616,738 |
| (b) | **11,806**  | 15,977 | 18,895  | 22,898  | 54,224  |
| (c) | **112.8**   | 31.0   | 34.7    | 46.8    | 118.9   |
| (d) | **10**      | 6      | 6       | 6       | 9       |
| (e) | **2.54**    | 5.45   | 4.65    | 5.21    | -       |

Table 1: Descriptive statistics: (a) total number of sentences, (b) vocabulary size, (c) average term frequency, (d) median number of words in a sentence, and (e) average number of shots.

| Noun | man, woman, girl, hand, hair, head, cat, boy, person |
|------|-------------------------------------------------------|
| Verb | be, look, wear, walk, dance, talk, smile, hold, sit  |
| Adj. | young, black, other, white, long, red, blond, dark  |

Table 2: Top frequent nouns/verbs/adjectives

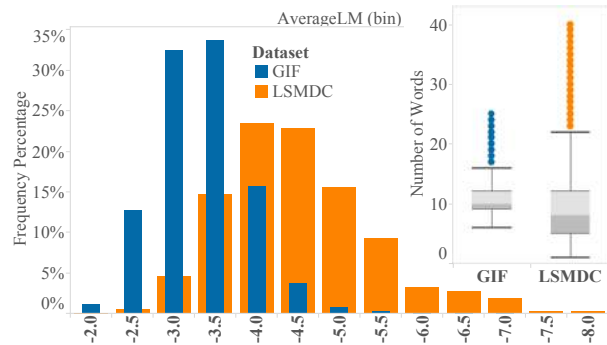

Figure 3: The main plot shows the distribution of language model scores averaged by the number of words in each dataset. The box plot shows the distribution of sentence lengths.

datasets are 6.13 and 3.02 seconds long, respectively.

Table 1 shows descriptive statistics of our dataset and existing datasets, and Table 2 shows the most frequent nouns, verbs and adjectives. Our dataset has more sentences with a smaller vocabulary size. Notably, our dataset has an average term frequency that is 3 to 4 times higher than other datasets. A higher average term frequency means less polymorphism, thus increasing the chances of learning visual/textual associations, an ideal property for image and video description.

**Language generality-specificity.** Our dataset is annotated by crowdworkers, while the movie datasets are annotated by trained professionals. As a result, the language in our dataset tends to be more general than the movie datasets. To show this, we measure how sentences in each dataset conform to "common language" using an n-gram language model (LM) trained on the Google 1B word corpus [4]. We average the LM score by the number of words in each sentence to avoid the tendency of a longer sentence producing a lower score. Figure 3 shows that our dataset has higher average LM scores even with longer sentences.

| Category | motion | contact | body | percp. | comm. |
|---|---|---|---|---|---|
| Examples | turn<br>move<br>walk<br>dance<br>shake | sit<br>stand<br>kiss<br>put<br>open | wear<br>smile<br>laugh<br>blow<br>dress | look<br>show<br>stare<br>watch<br>see | talk<br>wave<br>speak<br>point<br>nod |
| TGIF | **30%** | 17% | **11%** | 8% | **7%** |
| LSMDC | **31%** | 21% | 3% | **12%** | 4% |
| COCO | 19% | **35%** | 3% | 7% | 2% |

Table 3: Top verb categories with most common verbs for each category, and the distribution of verb occurrences on three datasets. Bold faced numbers are discussed in the text. "percp." means "perception", and "comm." means "communication."

|  | TGIF | LSMDC |
|---|---|---|
| *Q1*: Video contains a cohesive, self-contained visual story without any frame irrelevant to the main story. | 100.0%<br>±0.2% | 92.0%<br>±1.7% |
| *Q2*: Sentence accurately describes the main visual story of the video without missing information. | 95.0%<br>±1.4% | 78.0%<br>±2.6% |
| *Q3*: Sentence describes visual content available only within the video. | 94.0%<br>±1.5% | 88.0%<br>±2.0% |

Table 4: Polling results comparing TGIF and LSMDC datasets.

**Verb characteristics.** Identifying verbs (actions) is perhaps one of the most challenging problems in image and video description. In order to understand what types of verbs are used for describing each dataset, we link verbs in each sentence to WordNet using the semantic parser from [31]. Table 3 shows the distribution of top verb categories in each dataset (verb categories refer to the highest-level nodes in the WordNet hierarchy).

Not surprisingly, the MS-COCO dataset contains more static verbs (*contact*) compared to the video description datasets, which have more dynamic verbs (*motion*). This suggests that video contains more temporally dynamic content than static images. Most importantly, our dataset has more "picturable" verbs related to human interactions (*body*), and fewer abstract verbs (*perception*) compared to the LSMDC dataset. Because picturable verbs are arguably more visually identifiable than abstract verbs (e.g., walk vs. think), this result suggests that our dataset may provide an ideal testbed for video description.

**Quality of segmentation and description.** To make qualitative comparisons between the TGIF and LSMDC datasets, we conducted a user study designed to evaluate the quality of segmentation and language descriptions (see Table 4). The first question evaluates how well a video is segmented, while the other two evaluate the quality of text descriptions (how well a sentence describes the corresponding video). In the questionnaire we provided detailed examples for each question to facilitate complete understanding of the questions. We randomly selected 100 samples from each dataset, converted movie clips to animated GIFs, and mixed them in a random order to make them indistinguishable. We recruited 10 people from various backgrounds, and used majority voting to pool the answers from raters.

Table 4 shows two advantages of ours over LSMDC: (1) the animated GIFs are carefully segmented to convey a cohesive and self-contained visual story; and (2) the sentences are well associated with the main visual story.

## 5. Benchmark Evaluation

We report results on our dataset using three popular techniques used in video description: nearest neighbor, statistical machine translation, and LSTM.

### 5.1. Evaluation Metrics

We report performance on four metrics often used in machine translation: BLEU [27], METEOR [18], ROUGE [19] and CIDEr [37]. BLEU, ROUGE and CIDEr use only exact n-gram matches, while METEOR uses synonyms and paraphrases in addition to exact n-gram matches. BLEU is precision-based, while ROUGE is recall-based. CIDEr optimizes a set of weights on the TF-IDF match score using human judgments. METEOR uses an $F_1$ score to combine different matching scores. For all four metrics, a larger score means better performance.

### 5.2. Baseline Methods

The TGIF dataset is randomly split into 80K, 10K and 10K for training, validation and testing, respectively. The automatic animated GIF description methods learn from the training set, and are evaluated on the testing set.

#### 5.2.1 Nearest Neighbor (NN)

We find a nearest neighbor in the training set based on its visual representation, and use its sentence as the prediction result. Each animated GIF is represented using the off-the-shelf Hybrid CNN [46] and C3D [36] models; the former encodes static objects and scenes, while the latter encodes dynamic actions and events. From each animated GIF, we sample one random frame for the Hybrid CNN features and 16 random sequential frames for the C3D features. We then concatenate the two feature representations and determine the most similar instance based on the Euclidean distance.

#### 5.2.2 Statistical Machine Translation (SMT)

Similar to the two-step process of Rohrbach *et al*. [31], we automatically label an animated GIF with a set of semantic roles using a visual classifier and translate them into a sentence using SMT. We first obtain the semantic roles of words in our training examples by applying a semantic parser [31, 7]. We then train a visual classifier using the same input features as in the *NN* baseline and the semantic roles as the target variable. We use the multi-label classification model of Read *et al.* [29] as our visual classifier.

We compare two different databases to represent semantic roles: WordNet [22] and FrameNet [2], which we refer to as *SMT-WordNet* and *SMT-FrameNet*, respectively. For *SMT-WordNet*, we use the same semantic parser of Rohrbach *et al.* [31] to map the words into WordNet entries (semantic roles), while for *SMT-FrameNet* we use a frame semantic parser from Das *et al.* [7]. We use the phrase based model from Koehn *et al.* [15] to learn the SMT model.

### 5.2.3 Long Short-Term Memory (LSTM)

We evaluate an LSTM approach using the same setup of S2VT [38]. We also evaluate a number of its variants in order to analyze effects of different components.

**Basic setup.** We sample frames at 10 FPS and encode each using a CNN [16]. We then encode the whole sequence using an LSTM. After the encoding stage, a decoder LSTM is initialized with a "BOS" (Beginning of Sentence) token and the hidden states/memory cells from the last encoder LSTM unit. The decoder LSTM generates a description word-by-word using the previous hidden states and words, until a "EOS" (End of Sentence) token is generated. The model weights – CNN and encoder/decoder LSTM – are learned by minimizing the softmax loss $L$:

$$L = -\frac{1}{N} \sum_{i=1}^{N} \sum_{t=1}^{T} p(y_t = S_t^i | h_{t-1}, S_{t-1}^i), \quad (1)$$

where $S_t^i$ is the $t^{th}$ word of the $i^{th}$ sentence in the training data, $h_t$ is the hidden state and $y_t$ is the predicted word at timestamp $t$. The word probability $p(y_t = w)$ is computed as the softmax of the decoder LSTM output.

At the test phase, the decoder has no ground-truth word from which to infer the next word. There are many inference algorithms for this situation, including greedy, sampling, and beam search. We empirically found that the simple greedy algorithm performs the best. Thus, we use the most likely word at each time step to predict the next word (along with the hidden states).

We implemented the system using Caffe [13] on three K80 GPU cards, with the batch size fixed to 16, the learning rate decreasing from 0.1 to 1e-4 gradually, and for 16 epochs (800K iterations) over the training data. The optimization converges at around 600K iterations.

**Variants on cropping scheme.** We evaluate five variants of cropping schemes for data augmentation. S2VT uses a well-adopted spatial cropping [16] for all frames independently. To verify the importance of sequence modeling, we test *Single* cropping, where we take a single random frame from the entire sequence. *No-SP* crops 10 patches (2 mirrors of center, bottom/top-left/right) from each frame and average their CNN features. Spatial cropping is shown to be crucial to achieve a translation invariance for image recognition [33]. To achieve similar invariance effect along the temporal axis, we introduce *Tempo*, where a subsequence is randomly cropped from the original sequence and used as input for the sequence encoder (instead of the original full sequence); the spatial cropping is also applied to this baseline. S2VT crops patches from different spatial locations across frames. However, this introduces a spatial inconsistency into the LSTM encoder because the cropped location changes over the temporal axis. This may make it difficult to learn the right spatial-temporal dynamic to capture the motion information. Therefore, we introduce *Cubic* cropping, which adds a spatial consistency constraint to the *Tempo* version (see Figure 4).

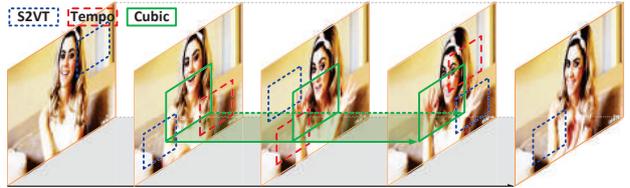

Figure 4: Illustration of three cropping schemes. S2VT crops patches from random locations across all frames in a sequence. The *Tempo* also crops patches from random locations, but from a randomly cropped subsequence. The *Cubic* crops patches from a random location shared across a randomly cropped subsequence.

**Variants on CNN weight optimization.** We evaluate three variants on how the CNN weights are initialized and updated. S2VT sets the weights by pretraining it on ImageNet-1K class categories [33] and fixing them throughout. The *Rand* model randomly initializes the CNN weights and fixes them throughout. To keep the CNN weights fixed, we limit the gradients of the loss function backpropagate only to the encoder LSTM. *Finetune* takes the pretrained parameters and finetunes them by backpropagating the gradients all the way down to the CNN part.

### 5.3. Results and Discussion

Table 5 summarizes the results. We can see that *NN* performs significantly worse than all other methods across all metrics. The *NN* copies sentences from the training set; our result suggests the importance of explicitly modeling sequence structure in GIFs and sentences for TGIF dataset.

**SMT baselines.** Our results show that *SMT-FrameNet* outperforms *SMT-WordNet* across the board. Does it mean the former should be preferred over the latter? To answer this, we dissect the two-step process of the SMT baseline by analyzing visual classification (image to semantic role) and machine translation (semantic role to sentence) separately. The mean $F_1$ score of visual classification on the test set is only 0.22% for WordNet; for FrameNet it is 2.09%. We also observe poor grammar performance with both variants, as is shown in Figure 5. We believe poor performance of visual classifiers has contributed to the poor grammar in generated sentences. This is because it makes the distribution of the input to the SMT system inconsistent with the training data. Although nowhere close to the current state-of-the-art image classification performance [33], the difference in mean $F_1$ scores in part explains the better performance of SMT-FrameNet, i.e., the second step (machine translation) receives more accurate classification results as input. We note, however, that there are 6,609 concepts from WordNet that overlaps with our dataset, while for FrameNet there are only 696 concepts. So the performance difference could merely reflect the difficulty of learning a visual classifier for WordNet with about 10 times more label categories.

| Methods | | BLEU-{1,2,3,4} | | | | METEOR | ROUGE_L | CIDEr |
|---|---|---|---|---|---|---|---|---|
| Nearest Neighbor | | 25.3 | 7.6 | 2.2 | 0.7 | 7.0 | 21.0 | 1.5 |
| SMT | WordNet | 27.8 | 13.6 | 6.3 | 3.0 | 9.6 | 26.0 | 8.9 |
| | FrameNet | 34.3 | 18.1 | 9.2 | 4.6 | 14.1 | 28.3 | 10.3 |
| LSTM | S2VT | 51.1 | 31.3 | 19.1 | 11.2 | 16.1 | 38.7 | 27.6 |
| | Crop Single | 47.0 | 27.1 | 15.7 | 9.0 | 15.4 | 36.9 | 23.8 |
| | Crop No-SP | 51.4 | 32.1 | 20.1 | 11.8 | 16.1 | 39.1 | 28.3 |
| | Crop Tempo | 49.4 | 30.4 | 18.6 | 10.6 | 16.1 | 38.4 | 26.7 |
| | Crop Cubic | 50.9 | 31.5 | 19.3 | 11.1 | 16.2 | 38.7 | 27.6 |
| | CNN Rand | 49.7 | 27.2 | 14.5 | 5.2 | 13.6 | 36.6 | 7.6 |
| | CNN **Finetune** | **52.1** | **33.0** | **20.9** | **12.7** | **16.7** | **39.8** | **31.6** |

Table 5: Benchmark results on three baseline methods and their variants on five different evaluation metrics.

We find a more conclusive answer by analyzing the machine translation step alone: We bypass the visual classification step by using ground-truth semantic roles as input to machine translation. We observe an opposite result: a METEOR score of 21.9% for SMT-FrameNet and 29.3% for SMT-WordNet. This suggests: (1) having a more expressive and larger semantic role vocabulary helps improve performance; and (2) there is huge potential for improvement on SMT-WordNet, perhaps more so than SMT-FrameNet, by improving visual classification of WordNet categories.

**LSTM baselines.** The LSTM methods significantly outperform the NN and the SMT baselines even with the simple CNN features – NN and SMT baselines use Hybrid CNN and C3D features. This conforms to recent findings that end-to-end sequence learning using deep neural nets outperforms traditional hand-crafted pipelines [43, 38]. By comparing results of different LSTM variants we make three major observations: (1) The fact that *Single* performs worse than all other LSTM variants (except for Rand) suggests the importance of modeling input sequence structure; (2) The four variants on different cropping schemes (S2VT, No-SP, Tempo, Cubic) perform similarly to each other, suggesting spatial and temporal shift-invariance of the LSTM approaches to the input image sequence; (3) Among the three variants of different CNN weight initialization and update schemes (S2VT, Rand, Finetune), Finetune performs the best. This suggests the importance of having a task-dependent representation in the LSTM baseline.

| | 20% | 40% | 60% | 80% | 100% |
|---|---|---|---|---|---|
| S2VT | 15.0 | 15.5 | 15.7 | 16.1 | 16.1 |

Table 6: METEOR scores improve as we use more training data, but plateau after 80% of the training set.

**Qualitative analysis.** Figure 5 shows sentences generated using the three baselines and their METEOR scores. The NN appears to capture some parts of visual components (e.g., (c) "drops" and (d) "white" in Fig. 5), but almost always fails to generate a relevant sentence. On the other hand, the SMT-FrameNet appears to capture more detailed semantic roles (e.g., (a) "ball player" and (b) "pool of water"), but most sentences contain syntactic errors. Finally, the LSTM-Finetune generates quite relevant and grammatical sentences, but at times fail to capture detailed semantics (e.g., (c) "running through" and (f) "a group of people"). We provide more examples in the supplementary material.

**Do we need more training data?** Table 6 shows the METEOR score of S2VT on various portions of the training dataset (but on the same test set). Not surprisingly, the performance increases as we use more training data. We see, on the other hand, that the performance plateaus after 80%. We believe this shows our TGIF dataset is already at its capacity to challenge current state-of-the-art models.

**Importance of multiple references.** Table 7 shows the METEOR score of three baselines according to different numbers of reference sentences in our test set. We see a clear pattern of increasing performance as we use more references in evaluation. We believe this reflects the fact that there is no clear cut single sentence answer to image and video description, and that it suggests using more references will increase the *reliability* of evaluation results. We believe the score will eventually converge with more references; we plan to investigate this in the future.

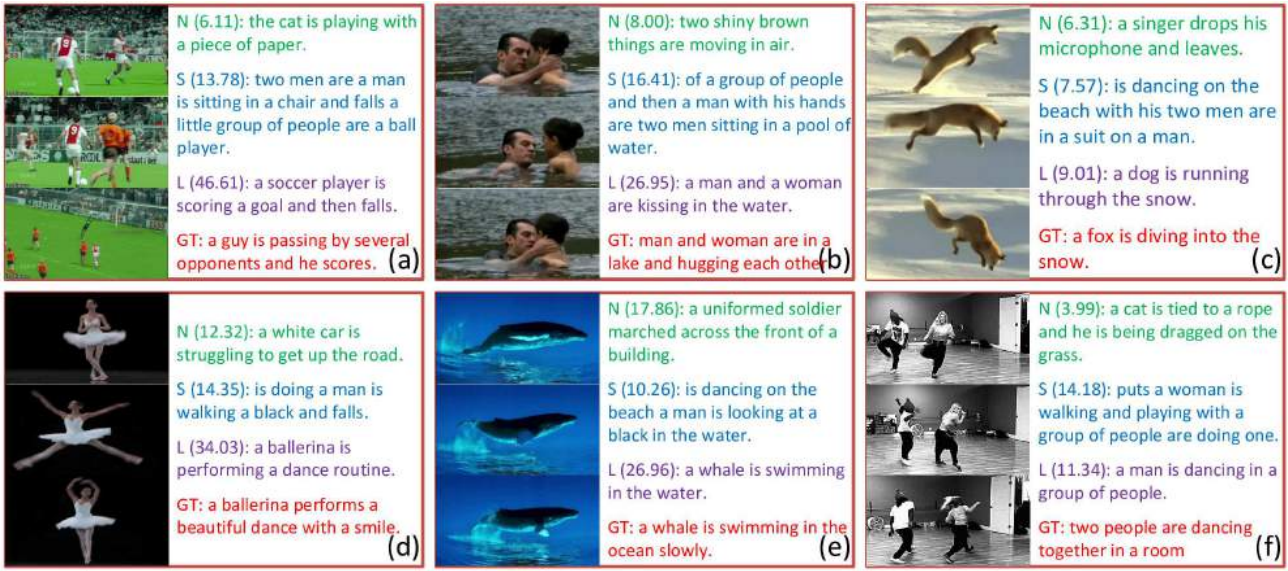

Figure 5: Example animated GIFs and generated sentences from nearest neighbor (N), SMT-FrameNet (S), and LSTM-Finetune (L). The GT refers to one of the 3 ground truth sentences provided by crowdworkers. The numbers in parentheses show the METEOR score (%) of each generated sentence. More examples can be found here: https://goo.gl/xcYjjE

| # of references | One | Two | Three |
|---|---|---|---|
| NN | 5.0 | 6.2 | 7.0 |
| SMT-FrameNet | 10.5 | 12.8 | 14.1 |
| LSTM-Finetune | 12.1 | 15.0 | 16.7 |

Table 7: METEOR scores improve with more reference sentences.

|  | M-VAD | MPII-MD | LSMDC |
|---|---|---|---|
| TGIF | 3.53 | 3.92 | 3.96 |
| Movie | 4.99 | 5.35 | **5.82** |
| TGIF-to-Movie | **5.17** | **5.42** | 5.77 |

Table 8: METEOR scores from cross-dataset experiments.

### 5.4. Cross-Dataset Adaptation: GIF to Movies

Finally, we evaluate whether an LSTM trained to describe animated GIFs can be applied to the movie description task. We test three settings (see Table 8). *TGIF* represents the basic S2VT model trained on the TGIF dataset, while *Movie* is the S2VT model trained on each movie dataset (M-VAD, MPII-MD, and LSMDC) respectively. Finally, *TGIF-to-Movie* represents the S2VT model pre-trained on the TGIF and fine-tuned on each of the movie datasets, respectively. We see that the *TGIF-to-Movie* improves performance on the M-VAD and MPII-MD datasets, and performs comparably to the LSMDC dataset.

### 6. Conclusions

We presented the Tumblr GIF (TGIF) dataset and showed how we solved multiple obstacles involved in crowdsourcing natural language descriptions, using automatic content filtering for collecting animated GIFs, as well as novel syntactic and semantic validation techniques to ensure high quality descriptions from free-form text input. We also provided extensive benchmark results using three popular video description techniques, and showed promising results on improving movie description using our dataset.

We believe TGIF shows much promise as a research tool for video description and beyond. An animated GIF is simply a limited series of still frames, often without narrative or need for context, and always without audio. So focusing on this constrained content is a more readily accessible bridge to advance research on video understanding than a leap to long-form videos, where the content is complex with contextual information that is currently far from decipherable automatically. Once the content of animated GIFs is more readily recognizable, the step to video understanding will be more achievable, through adding audio cues, context, storytelling archetypes and other building blocks.

### Acknowledgements

This work was supported in part by Yahoo Research, Flickr, and New York State through the Goergen Institute for Data Science at the University of Rochester. We thank Gerry Pesavento, Huy Nguyen and others from Flickr for their support in collecting descriptions via crowdsourcing.

## References

[1] S. Antol, A. Agrawal, J. Lu, M. Mitchell, D. Batra, C. L. Zitnick, and D. Parikh. VQA: Visual question answering. In *ICCV*, 2015. 2

[2] C. F. Baker, C. J. Fillmore, and J. B. Lowe. The berkeley framenet project. In *ACL*, 1998. 6

[3] S. Bakhshi, D. Shamma, L. Kennedy, Y. Song, P. de Juan, and J. J. Kaye. Fast, cheap, and good: Why animated GIFs engage us. In *CHI*, 2016. 1

[4] C. Chelba, T. Mikolov, M. Schuster, Q. Ge, T. Brants, P. Koehn, and T. Robinson. One billion word benchmark for measuring progress in statistical language modeling. In *INTERSPEECH*, 2014. 4

[5] D. L. Chen and W. B. Dolan. Collecting highly parallel data for paraphrase evaluation. In *ACL*, 2011. 2, 4

[6] J. Daiber, M. Jakob, C. Hokamp, and P. N. Mendes. Improving efficiency and accuracy in multilingual entity extraction. In *I-Semantics*, 2013. 3

[7] D. Das, D. Chen, A. F. Martins, N. Schneider, and N. A. Smith. Frame-semantic parsing. *Computational Linguistics*, 40(1), 2014. 5, 6

[8] P. Das, C. Xu, R. Doell, and J. Corso. A thousand frames in just a few words: Lingual description of videos through latent topics and sparse object stitching. In *CVPR*, 2013. 2

[9] H. Fang, S. Gupta, F. Iandola, R. K. Srivastava, L. Deng, P. Dollár, J. Gao, X. He, M. Mitchell, J. C. Platt, et al. From captions to visual concepts and back. In *CVPR*, 2015. 2

[10] A. Farhadi, M. Hejrati, M. A. Sadeghi, P. Young, C. Rashtchian, J. Hockenmaier, and D. Forsyth. Every picture tells a story: Generating sentences from images. In *ECCV*. 2010. 2

[11] S. Guadarrama, N. Krishnamoorthy, G. Malkarnenkar, S. Venugopalan, R. Mooney, T. Darrell, and K. Saenko. Youtube2text: Recognizing and describing arbitrary activities using semantic hierarchies and zero-shot recognition. In *ICCV*, 2013. 2

[12] S. Hochreiter and J. Schmidhuber. Long short-term memory. *Neural computation*, 9(8), 1997. 2

[13] Y. Jia, E. Shelhamer, J. Donahue, S. Karayev, J. Long, R. Girshick, S. Guadarrama, and T. Darrell. Caffe: Convolutional architecture for fast feature embedding. In *MM*, 2014. 6

[14] A. Karpathy and L. Fei-Fei. Deep visual-semantic alignments for generating image descriptions. In *CVPR*, 2015. 2

[15] P. Koehn, H. Hoang, A. Birch, C. Callison-Burch, M. Federico, N. Bertoldi, B. Cowan, W. Shen, C. Moran, R. Zens, et al. Moses: Open source toolkit for statistical machine translation. In *ACL*, 2007. 6

[16] A. Krizhevsky, I. Sutskever, and G. E. Hinton. Imagenet classification with deep convolutional neural networks. In *NIPS*, 2012. 6

[17] G. Kulkarni, V. Premraj, V. Ordonez, S. Dhar, S. Li, Y. Choi, A. C. Berg, and T. Berg. Babytalk: Understanding and generating simple image descriptions. *PAMI*, 35(12), 2013. 2

[18] M. D. A. Lavie. Meteor universal: Language specific translation evaluation for any target language. *ACL*, 2014. 4, 5

[19] C.-Y. Lin. Rouge: A package for automatic evaluation of summaries. In *ACL*, 2004. 5

[20] T. Lin, M. Maire, S. J. Belongie, J. Hays, P. Perona, D. Ramanan, P. Dollár, and C. L. Zitnick. Microsoft COCO: common objects in context. In *ECCV*, 2014. 1, 2, 4

[21] J. Mao, W. Xu, Y. Yang, J. Wang, and A. Yuille. Deep captioning with multimodal recurrent neural networks (m-rnn). *arXiv preprint arXiv:1412.6632*, 2014. 2

[22] G. A. Miller. Wordnet: a lexical database for english. *Communications of the ACM*, 38(11), 1995. 6

[23] L. Neumann and J. Matas. Real-time scene text localization and recognition. In *CVPR*, 2012. 3

[24] M. Norouzi, A. Punjani, and D. J. Fleet. Fast exact search in hamming space with multi-index hashing. *PAMI*, 36(6), 2014. 3

[25] V. Ordonez, G. Kulkarni, and T. L. Berg. Im2text: Describing images using 1 million captioned photographs. In *NIPS*, 2011. 2

[26] Y. Pan, T. Mei, T. Yao, H. Li, and Y. Rui. Jointly modeling embedding and translation to bridge video and language. *CoRR*, abs/1505.01861, 2015. 2

[27] K. Papineni, S. Roukos, T. Ward, and W.-J. Zhu. Bleu: a method for automatic evaluation of machine translation. In *ACL*, 2002. 5

[28] C. Rashtchian, P. Young, M. Hodosh, and J. Hockenmaier. Collecting image annotations using amazon's mechanical turk. In *NAACL HLT*, 2010. 4

[29] J. Read, B. Pfahringer, G. Holmes, and E. Frank. Classifier chains for multi-label classification. *Machine learning*, 85(3), 2011. 6

[30] M. Regneri, M. Rohrbach, D. Wetzel, S. Thater, B. Schiele, and M. Pinkal. Grounding action descriptions in videos. *TACL*, 2013. 2

[31] A. Rohrbach, M. Rohrbach, N. Tandon, and B. Schiele. A dataset for movie description. In *CVPR*, 2015. 1, 2, 3, 4, 5, 6

[32] M. Rohrbach, W. Qiu, I. Titov, S. Thater, M. Pinkal, and B. Schiele. Translating video content to natural language descriptions. In *ICCV*, 2013. 2

[33] O. Russakovsky, J. Deng, H. Su, J. Krause, S. Satheesh, S. Ma, Z. Huang, A. Karpathy, A. Khosla, M. Bernstein, A. C. Berg, and L. Fei-Fei. ImageNet Large Scale Visual Recognition Challenge. *IJCV*, 2015. 1, 6, 7

[34] A. Torabi, P. Chris, L. Hugo, and C. Aaron. Using descriptive video services to create a large data source for video annotation research. *arXiv preprint*, 2015. 1, 2, 4

[35] K. Toutanova and C. D. Manning. Enriching the knowledge sources used in a maximum entropy part-of-speech tagger. In *ACL*, 2010. 3

[36] D. Tran, L. D. Bourdev, R. Fergus, L. Torresani, and M. Paluri. Learning spatiotemporal features with 3d convolutional networks. In *ICCV*, 2015. 3, 5

[37] R. Vedantam, C. Lawrence Zitnick, and D. Parikh. Cider: Consensus-based image description evaluation. In *CVPR*, 2015. 5

[38] S. Venugopalan, M. Rohrbach, J. Donahue, R. Mooney, T. Darrell, and K. Saenko. Sequence to sequence - video to text. In *ICCV*, 2015. 2, 6, 7

[39] S. Venugopalan, H. Xu, J. Donahue, M. Rohrbach, R. J. Mooney, and K. Saenko. Translating videos to natural language using deep recurrent neural networks. In *NAACL HLT*, 2015. 2

[40] O. Vinyals, A. Toshev, S. Bengio, and D. Erhan. Show and tell: A neural image caption generator. In *CVPR*, 2015. 2

[41] K. Xu, J. Ba, R. Kiros, K. Cho, A. C. Courville, R. Salakhutdinov, R. S. Zemel, and Y. Bengio. Show, attend and tell: Neural image caption generation with visual attention. In *ICML*, 2015. 2

[42] B. Z. Yao, X. Yang, L. Lin, M. W. Lee, and S.-C. Zhu. I2t: Image parsing to text description. *IEEE*, 1998. 2

[43] L. Yao, A. Torabi, K. Cho, N. Ballas, C. Pal, H. Larochelle, and A. Courville. Describing videos by exploiting temporal structure. In *ICCV*, 2015. 2, 7

[44] L. Yu, E. Park, A. C. Berg, and T. L. Berg. Visual madlibs: Fill in the blank description generation and question answering. In *ICCV*, 2015. 2

[45] C. Zauner. *Implementation and benchmarking of perceptual image hash functions*. 2010. 3

[46] B. Zhou, A. Lapedriza, J. Xiao, A. Torralba, and A. Oliva. Learning deep features for scene recognition using places database. In *NIPS*, 2014. 5

# Supplementary Material for
## *TGIF: A New Dataset and Benchmark on Animated GIF Description*

Yuncheng Li  
University of Rochester

Yale Song  
Yahoo Research

Liangliang Cao  
Yahoo Research

Joel Tetreault  
Yahoo Research

Larry Goldberg  
Yahoo Research

Alejandro Jaimes  
AiCure

Jiebo Luo  
University of Rochester

This supplementary material contains detailed information about the paper, specifically:

- Figure 1: A full version of the instructions shown to the crowd workers iIn reference to Section 3.2. Data Annotation).

- Figure 2: The task page shown to the workers (in reference to Section 3.2. Data Annotation).

- Figure 3: Syntactic validation error messages shown to the workers (In reference to Section 3.2).

- Figure 4: Example animated GIFs with generated descriptions (in reference to Section 5.3 Results and Discussion).

- Figure 5: Side-by-side qualitative comparisons of sentences provided in the LSMDC dataset (professionally annotated descriptive video service (DVS) captions) and sentences annotated by the crowd workers (in reference to Section 2.1. Comparison with LSMDC).

In Figure 4 and Figure 5, we provide one still frame of each animated GIF for easy visualization. However, the still frames are inadequate for showing the dynamic visual content of animated GIFs. We therefore invite the reviewers to the following websites as an alternative to Figure 4 and Figure 5. Each website also contains more examples than we provide in this supplementary material.

- Figure 4: `https://goo.gl/xcYjjE`

- Figure 5: `https://goo.gl/ZGYIYh`

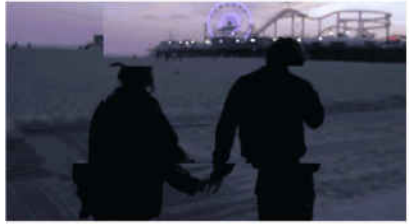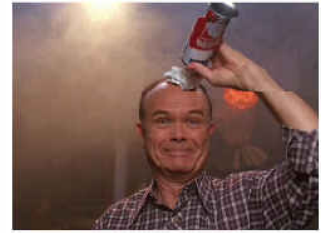

Figure 1: A full version of the instructions shown to the crowd workers.

Figure 2: **Task page.** Each task page showed 5 animated GIFs and asked the worker to describe each with one sentence. This figure shows one of the five questions in a task page.

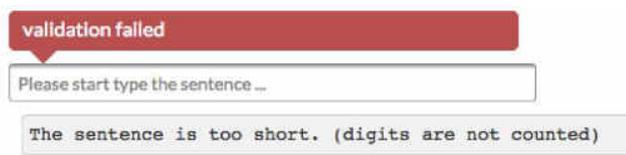

(a) A sentence must contain at least 8, but no more than 25 words.

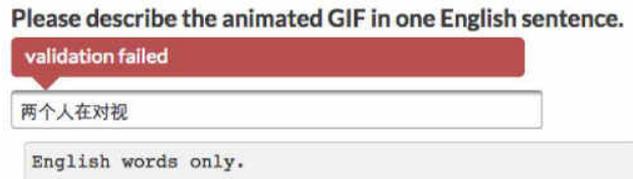

(b) A sentence must contain only English words.

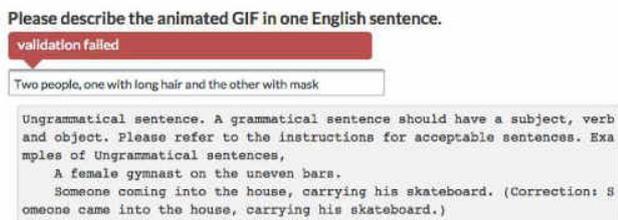

(c) A sentence must contain a main verb.

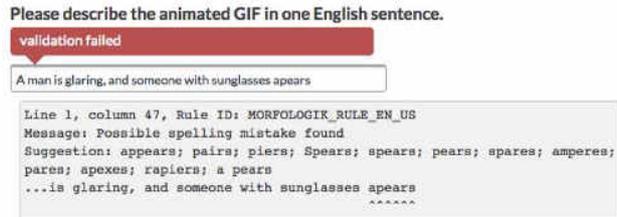

(d) A sentence must be grammatical and free of typos.

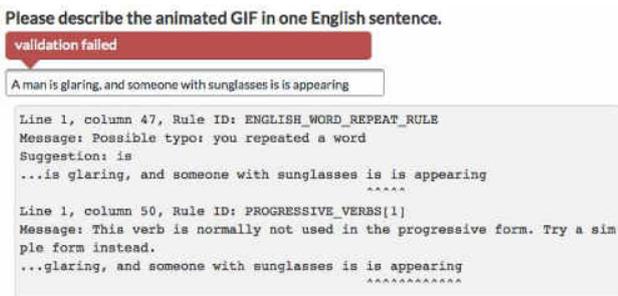

(e) A sentence must be grammatical and free of typos.

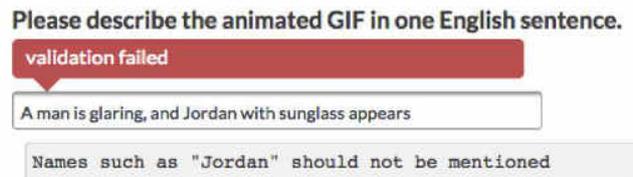

(f) A sentence must contain no named entities, such as an actor/actress, movie names, countries.

Figure 3: **Syntactic validation.** Since the workers provide free-form text, we validate the sentence before submission to reduce syntactic errors. This figure shows example messages shown to the workers in case we detect a syntactic error: error type (red speech balloon) and explanation for the error (caption below text input).

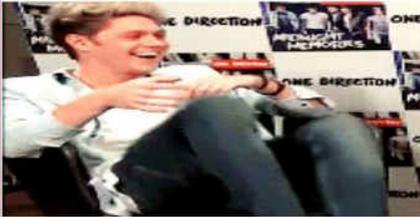
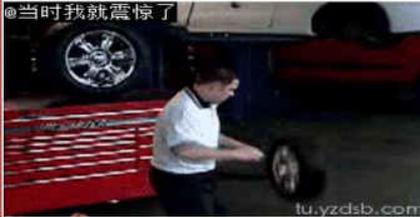
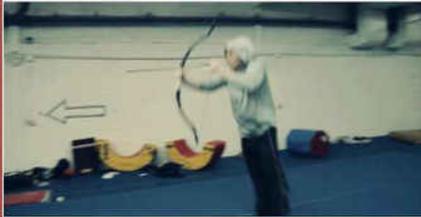
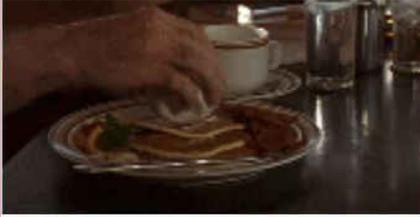
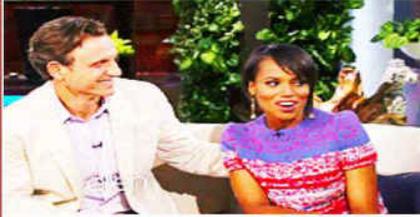
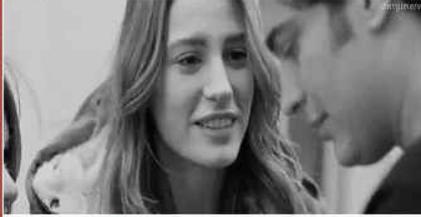
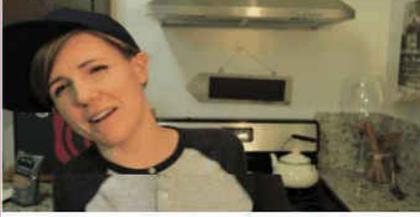
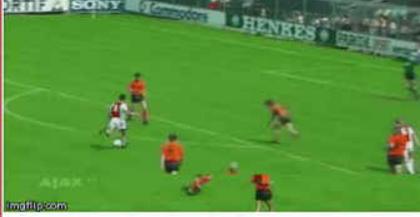
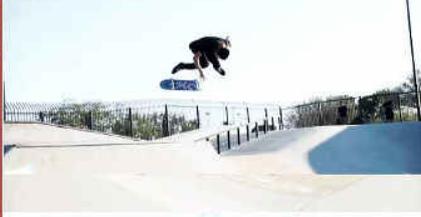
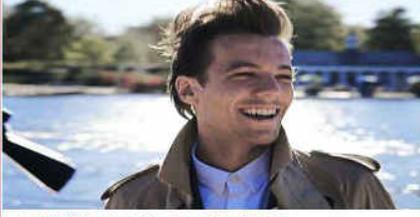
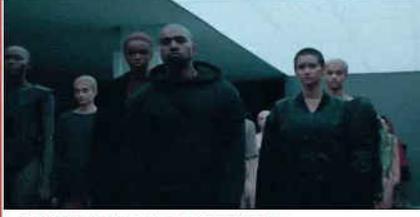
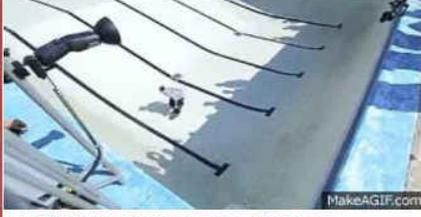

Figure 4: **Example animated GIFs and sentences generated** using the following algorithms: nearest neighbor (N), SMT-FrameNet (S), and LSTM-Finetune (L). The numbers in parentheses show the METEOR score (%) of each generated sentence. More examples with better visualization: https://goo.gl/xcYjjE

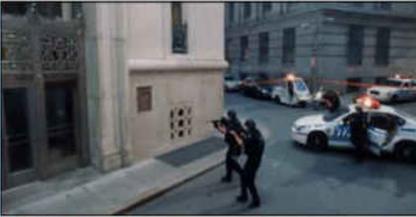

Figure 5: **Side-by-side qualitative comparisons of sentences provided in the LSMDC dataset (professionally annotated descriptive video service (DVS) captions) and sentences annotated by the crowd workers.** We converted movie clips from the LSMDC dataset into the GIF format and used the same pipeline described in our paper (Section 3) to crowdsource descriptions for the clips. Below each movie clip, we show both the crowdsourced description (CW) and the original description found in the LSMDC dataset (MD). We can see that the DVS captions tend to focus on visual content related specifically to the story (e.g., example No.6) or rely on contextual information not actually present within a video clip (e.g., "bear" in example No.11). More examples with better visualization: https://goo.gl/ZGYIYh